\documentclass[]{fairmeta}

\usepackage{microtype}
\usepackage{graphicx}
\usepackage{booktabs} 
\usepackage{adjustbox}

\usepackage{hyperref}
\usepackage{arydshln}

\usepackage{amsmath}
\usepackage{amssymb}
\usepackage{mathtools}
\usepackage{amsthm}
\usepackage{multicol}
\usepackage{multirow}
\usepackage{resizegather}

\theoremstyle{plain}

\theoremstyle{definition}

\theoremstyle{remark}

\newtheorem{example}{Example}

\usepackage[textsize=tiny]{todonotes}

\newcommand{\methodname}{VToM}

\title{Through the Theory of Mind's Eye: Reading Minds with Multimodal Video Large Language Models}

\author[1]{Zhanwen Chen}
\author[2]{Tianchun Wang}
\author[3]{Yizhou Wang}
\author[4]{Michal Kosinski}
\author[2]{Xiang Zhang}
\author[3]{Yun Fu}
\author[1]{Sheng Li}

\affiliation[1]{University of Virginia}
\affiliation[2]{The Pennsylvania State University}
\affiliation[3]{Northeastern University}
\affiliation[4]{Stanford University}

\abstract{
Recent work has revealed that large language models (LLMs) can exhibit emergent theory-of-mind (ToM) capabilities—inferring human beliefs, desires, and intentions from text alone. Yet, everyday social reasoning often unfolds visually in dynamic contexts. This paper investigates whether multimodal LLMs can similarly demonstrate ToM skills in video-based tasks. Concretely, we propose a pipeline that fuses video and text signals, retrieves the most relevant frames for each query, and answers questions requiring spatio-temporal social understanding. We introduce a new frame localization benchmark, Theory of Mind Localization (ToMLoc), and show that finetuning a state-of-the-art Video-ChatGPT model on ToMLoc significantly improves performance on Social-IQ 2.0. Our results suggest that bridging textual and visual modalities is essential for capturing complex mental states in real-world scenarios. Moreover, retrieving key frames enhances interpretability by revealing how the model arrives at its inferences. These findings highlight the promise of video-based approaches for achieving more human-like social intelligence in LLMs.
}

\date{\today}
\correspondence{Zhanwen Chen at \email{pct4et@virginia.edu}}
\code{\href{https://github.com/zhanwenchen/vtom}{https://github.com/zhanwenchen/vtom}}
\data{\href{https://huggingface.co/datasets/zhanwenchen/vtom_bench}{https://huggingface.co/datasets/zhanwenchen/vtom\_bench}}

\begin{document}
\maketitle

\vspace{-0.1cm}
\section{Introduction}
\vspace{-0.1cm}
Recent research has discovered emergent theory-of-mind (ToM) reasoning abilities in text-based large language models (LLMs) such as ChatGPT-4~\cite{kosinski2024evaluating, strachan2024}.
This finding is surprising in many respects. For the first time, an artificial intelligence model appears able to \textit{simulate} the act of ‘reading minds,’ i.e., inferring beliefs, desires, and intentions purely from textual clues. Consider a classic false-belief scenario: John pranks Kelly by placing marbles inside a chocolate box that he gives her for her birthday. A ToM-capable model would predict that Kelly will be \textit{surprised} upon opening the box, believing it to contain chocolates rather than marbles. Although recent work suggests that large LLMs can, in many instances, make the correct inference, subsequent studies have revealed potential limitations to these emergent ToM-like capabilities.

Indeed, the supporting evidence remains mixed. Some reports indicate that if a box is made transparent—thereby removing the possibility of a mistaken belief—LLMs may fail to incorporate this new visual caveat and still conclude that Kelly has a false belief. Yet, other findings~\cite{kosinski2024evaluating} suggest that more advanced models, such as GPT-4, often navigate these “control tasks” more deftly, implying that the exact nature and consistency of their ToM reasoning abilities remain under active investigation. Consequently, questions persist as to whether LLMs are systematically encoding deeper mental-state reasoning or relying on statistical shortcuts driven by textual prompts.

A more fundamental question emerges when we consider that human social reasoning is inherently multimodal, dynamically combining linguistic, visual, and contextual cues over time. For instance, individuals effortlessly read facial expressions, track eye movements, and interpret subtle gestures that may span several seconds or minutes. By contrast, many existing LLM-based ToM tests are static, text-only tasks. While these can reveal interesting linguistic and logical competencies, they do not fully capture the richness of how humans actually infer the contents of others’ minds in real-world situations.

To address these gaps, we argue for a **video-based** ToM approach. Videos present a temporal dimension that naturally lends itself to capturing how beliefs, desires, and intentions unfold and interact. For example, a character who first sees an event might later revise their understanding after new evidence appears in the scene. Such chronological ordering is often critical to establishing genuine theory-of-mind reasoning, as mental states are rarely static. Meanwhile, recent work by~\cite{shain2022evidence} using fMRI suggests that the human language network does \textit{not} robustly encode ToM, indicating that distinct neurobiological subsystems handle mentalizing. This raises doubts as to whether purely text-trained language models can ever truly internalize the spatiotemporal richness of real human ToM.

Against this backdrop, our project targets the research challenge of improving theory-of-mind (ToM) reasoning in \textbf{video-based large language models}. Prior investigations of ToM in LLMs mostly focused on text-based tasks~\cite{kosinski2024evaluating}, but we posit that the essence of ToM—tracking mental states as they evolve—demands integrating video. Accordingly, we propose a novel \emph{video frame retrieval} task tailored to localize ToM understanding within specific segments. Rather than restricting the model to a single text passage, we ask it to identify and ground its inferences in the relevant visual frames or short clips that convey a character’s belief, emotion, or intention at a particular time.

\begin{figure*}[ht!]
    \centering
    \includegraphics[width=0.9\textwidth]{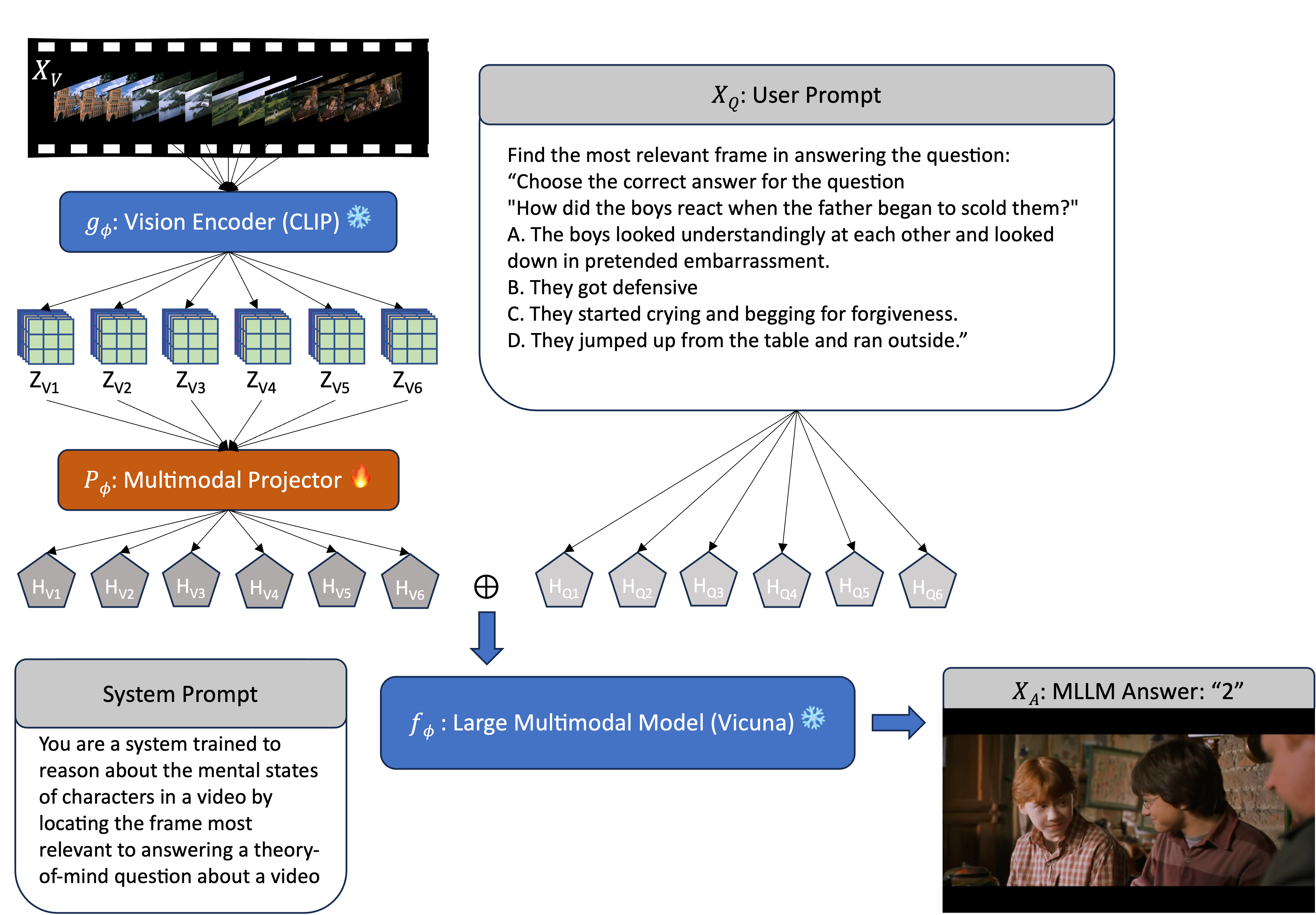}
    \caption{\textbf{An overview of our proposed Video Theory-of-Mind (VToM) architecture for solving Theory-of-Mind reasoning tasks in videos.}}
    \label{fig:pipeline}
\end{figure*}

This approach naturally leads to a \textit{multitask} formulation that encompasses both explicit \textbf{temporal ToM reasoning} and global question answering. Concretely, we develop a new architecture, \textbf{Video Theory of Mind (VToM)}, that captures the evolution of mental states. This module draws upon textual information from video transcripts and visual features from advanced video captioning models. Specifically, we leverage state-of-the-art “video LLMs,” exemplified by Video-ChatGPT~\cite{Maaz2023VideoChatGPT}, to fuse and interpret these distinct data streams. We then finetune the system on a ToM reasoning task. Figure~\ref{fig:pipeline} illustrates this overarching pipeline.

Furthermore, we evaluate our approach on a ToM-rich video dataset, \textbf{Social-IQ 2.0}~\cite{siq2}, where real-life YouTube interactions often involve intricate social and emotional cues. In these evaluations, we focus not only on overall question-answer accuracy but also on \emph{frame localization} metrics—whether the model can pinpoint precisely when and where key ToM-relevant events occur. We also release semi-supervised GPT-4V~\cite{openai2023gpt4v} annotations for further research on ToM labeling.

Beyond our technical objectives, we emphasize a broader motivation: if future LLMs genuinely capture aspects of ToM in dynamic social contexts, they may significantly influence user perceptions, trust, and engagement with AI systems. For instance, an AI assistant that can \emph{anticipate} user confusion or empathize with frustration could deliver more tailored, comforting, or motivational responses. Conversely, the power to track and predict mental states in real time could pose privacy and ethical dilemmas, raising the question of how to regulate such psychologically aware technologies. While our current work concentrates on the algorithmic and evaluative aspects of ToM in multimodal LLMs, we hope to spark further discourse on these social and psychological implications.

\textbf{In summary, we make the following key contributions:}
\begin{enumerate}
    \item \textbf{A new frame retrieval task for video ToM:} We propose \textit{Theory of Mind Localization (ToMLoc)}, a dataset and challenge focusing explicitly on temporal reasoning about mental states in video. We show that our pipeline establishes a 77.97\% frame localization baseline for future work.
    \item \textbf{Multimodal ToM architecture:} We introduce \emph{Video Theory of Mind (VToM)}, which combines textual transcripts and video features using advanced LLM architectures like Video-ChatGPT, enabling modeling of changing mental states over time.
    \item \textbf{Evaluation on Social-IQ 2.0:} We finetune and benchmark our approach on a ToM-centric subset of Social-IQ 2.0~\cite{siq2}. We demonstrate that the ToM-finetuned model significantly outperforms existing baselines, achieving a 47.15\% accuracy on the question-answering task.
    \item \textbf{GPT-4V labels for interpretability:} We release semi-supervised GPT-4V annotations as an additional benchmark resource, allowing future research to compare or augment human ToM labeling with LLM-generated insights.
\end{enumerate}

The remainder of this paper is organized as follows. Section~\ref{sec:related_work} provides background on theory-of-mind in psychology and computational approaches in both text and video. Section~\ref{sec:method} explains our ToMLoc task, data processing, and the VToM pipeline in detail. Section~\ref{sec:experiments} outlines our experimental setup and key results, discusses limitations, potential future directions, and ethical considerations of building ToM-capable AI systems.
\vspace{-0.1cm}
\section{Related Work} \label{sec:related_work}
\vspace{-0.1cm}

\textbf{Human Theory-of-Mind Reasoning on Text.}
Traditional Theory of Mind (ToM) Reasoning tasks and assessments revolve around text. In addition to text, we engage in ToM reasoning when we watch movies or television. In fact, we sometimes do so explicitly when we discuss films or TV shows with one another, such as discussing character intentions or the morality of certain actions in dynamic scenes. Indeed, video-based ToM reasoning can serve as an important developmental tool. \cite{valkenburg1999developing} found that parents often use videos as a form of \textit{instructive mediation} (also known as \textit{active mediation}), where they watch a video and ask their children about the motives that characters might have, or why certain actions are considered good or bad. The children's answers are accompanied by instructional feedback from the parents. Video-based instructive mediation is especially useful, considering that children can spend between 2.5 and 3 hours watching videos a day~\cite{rideout2019common}. In the context of developmental psychology, the intervention for autism spectrum disorder (ASD) incorporates many ToM tasks in formal assessments~\cite{heavey2000awkward, Dziobek2006}, neuroimaging stimulus materials~\cite{jacoby2016localizing}, behavioral studies~\cite{black2015fiction}, and interventions~\cite{Muller2017using}. A notable example is the ``Movie Time Social Learning'' program~\cite{vagin2012movie}, which uses movie discussions to assist children with autism in grasping social contexts and understanding others' points of view.

\textbf{Computational Theory-of-Mind Reasoning on Text.} Efforts on computational theory of mind encompass a diverse range of datasets aimed at enhancing AI's understanding and prediction of human mental states and social interactions. The Theory of Mind Task dataset provides brief text stories paired with ToM-focused questions, facilitating the evaluation of AI's ability to reason about mental states~\cite{nematzadeh2018evaluating}. The Visual Beliefs dataset offers short comic-style image sequences that challenge models to identify mistaken beliefs, a key aspect of theory of mind~\cite{eysenbach2016mistaken}. Additionally, the Motivations dataset includes images of individuals annotated with probable motivations, allowing models to infer underlying intentions~\cite{vondrick2016predicting}.

Narrative descriptions of prevalent social scripts and cultural norms further contribute to this domain by providing context-specific social knowledge~\cite{li2012learning}. Social Stories~\cite{Sap2019socialiqa} are specifically designed to help individuals with autism improve their interpersonal skills via better understanding and navigating social situations. The Story Commonsense and GLUCOSE datasets contain stories annotated with commonsense information, enabling models to grasp everyday logic and causality in narratives~\cite{Chen2019Incorporating, Mostafazadeh2020glucose}. Finally, the Neural Theory-of-Mind (TOMI) dataset focuses on assessing the ability of large language models to reason about mental states and social interactions, highlighting current limitations and future directions for research~\cite{sap-etal-2022-neural}.

These combined efforts are crucial for advancing the field of AI, making machines better at understanding and predicting human behavior and interactions.

\textbf{Theory-of-Mind Reasoning with Large Language Models.}~\cite{kosinski2024evaluating} examined the performance of large language models (LLMs) on text-based false-belief tasks and unexpected transfer tasks and found that the latest model, ChatGPT-4, performs at a level comparable to that of seven-year-old children on these tasks. The study suggests that LLMs' performance on these tasks is driven by a single factor, possibly an ability to detect false belief. The high correlation between LLMs' performance on both types of tasks suggests that a single factor, such as the ability to detect false belief, is driving their performance rather than separate task-specific abilities.



\vspace{-0.1cm}
\section{\methodname{}: \textbf{V}ideo \textbf{T}heory of \textbf{M}ind} \label{sec:method}
\vspace{-0.1cm}
In this section, we detail our proposed approach for improving theory-of-mind (ToM) reasoning in video-based large language models (LLMs). We first introduce the \emph{ToM Localization (ToMLoc)} task, designed to pinpoint frames that reveal mental-state cues, and then describe how we extend a baseline video LLM to handle this challenge.

\subsection{The ToM Localization (ToMLoc) Task}

Conventional video-text retrieval tasks typically seek to identify an entire clip that best matches a textual description. By contrast, \emph{ToMLoc} tackles the fine-grained problem of locating \emph{which frames} (or short segments) within a video are crucial for inferring a character’s beliefs, intentions, or emotions. Formally, given a video $X_{V}$ and a ToM-related question $X_{Q}$, the goal is to retrieve the subset of frames $X_{I}$ that most directly support answering $X_{Q}$.

An additional layer of complexity arises from the temporal nature of social interactions: mental states evolve over time and may only become apparent at certain moments. Thus, the ToMLoc task highlights where and \emph{when} a character’s attitude or understanding changes, rather than focusing solely on broad video-level matches.

\subsection{Example of ToM Frame Retrieval}

To illustrate the motivation and nature of the ToMLoc task, consider the following example scenario:

\begin{example}\label{th:example}
    \textbf{Scenario:} In the TinySocial dataset, Harry Potter visits the Weasleys’ home. The father begins scolding the boys, prompting the question: “How did the boys react when the father began to scold them?” Answer choices might include:
    (A) They exchanged knowing looks and pretended embarrassment,
    (B) They got defensive,
    (C) They begged for forgiveness,
    (D) They ran outside.
\end{example}

Answering this question requires detecting when the father starts scolding the boys, observing their body language (e.g., shared glances, lowered heads), and interpreting the emotional undertone (pretended embarrassment rather than genuine fear). This example highlights the need for fine-grained temporal localization: traditional retrieval might only match the phrase “father scolding” to a generic moment, ignoring the subtlety of what the boys \emph{think} or \emph{feel}. ToMLoc aims to localize frames where mental states are visible—such as a conspiratorial look or nervous fidgeting—so the model can base its answers on concrete visual cues.

\subsection{Frame Relevance Scoring and Temporal Reasoning}

To systematically identify key frames, we define a \emph{frame relevance scoring} mechanism. Let \( F = \{f_1, f_2, \ldots, f_n\} \) be the sequence of frames extracted from the video, and \( Q = \{q_1\} \) be a ToM-focused query. We compute:
\[
R(f_i, q_1) = \text{SIM} \bigl(\phi(f_i), \psi(q_1)\bigr),
\]
where \( \phi(\cdot) \) projects a frame into a visual embedding space, and \( \psi(\cdot) \) embeds the query into a textual (or multimodal) space. The similarity function, \( \text{SIM}(\cdot, \cdot) \), could be cosine similarity or another learned metric. Then, we select:
\[
F^* = \arg\max_{F' \subseteq F} \sum_{q_1 \in Q} \max_{f_i \in F'} R(f_i, q_1),
\]
which maximizes the overall alignment between the question and the candidate frames. In ToM-rich scenarios, this mechanism identifies frames containing facial expressions, gestures, or other cues revealing mental states key to answering $Q$.

\subsection{Extending a Baseline Video LLM}


Figure~\ref{fig:pipeline} illustrates how we incorporate ToMLoc into a baseline video LLM (e.g., Video-ChatGPT). First, a pretrained visual encoder $g_{\phi}$ extracts features $Z_{V}$ from video frames. A linear mapping $P$ then converts $Z_{V}$ into tokens $H_{V}$ compatible with the LLM’s embedding space. Simultaneously, a tokenizer $T$ processes the textual query $X_{Q}$ into tokens $H_{Q}$. Concatenating these embeddings as $(H_{V} \oplus H_{Q})$, we feed them into the model $f$ to produce an answer $X_{I}$ (or a ranked list of relevant frames). Formally:

\begin{equation}\label{eq:pipeline}
X_{I} = f\bigl(P(Z_{V}) \oplus T(X_{Q})\bigr).
\end{equation}

The model is then finetuned so that it (1) \emph{retrieves} the frames that best address ToM questions and (2) provides accurate answers about mental states. This strategy encourages the system to attend directly to those visual elements that reveal psychological nuance, rather than merely matching high-level keywords or contextual cues.

\subsection{Approximating Influential Frames with Median Frame}

\begin{figure*}[ht!]
    \centering
    \includegraphics[width=1.0\textwidth]{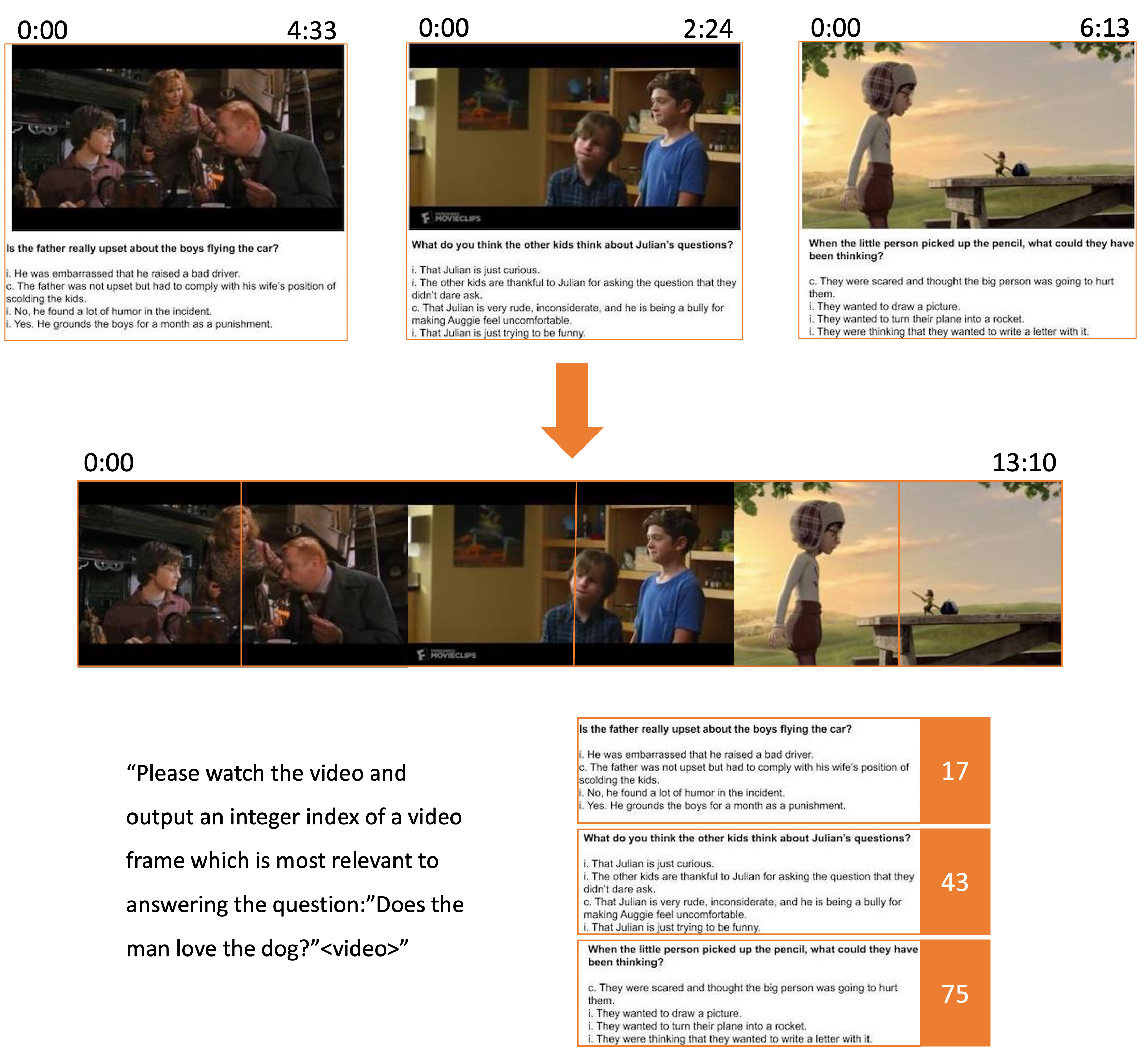}
    \caption{\textbf{The Median Frame Approximation Task for Most Influential Frame Retrieval.}}
    \label{fig:mf}
\end{figure*}

A practical challenge arises in datasets such as Social-IQ, where frame-level annotations do not exist. Human labeling of precisely which frames convey critical ToM information can be expensive and time-consuming. As a stopgap solution, we propose a \emph{median-frame approximation}, where each short clip is represented by its middle frame. We concatenate multiple clips into a single composite video and assume that any query about a given clip pertains to the region spanned by that clip. The median frame then serves as a heuristic “key moment” (Figure~\ref{fig:mf}).

For example, a composite video $V_{123}$ may include segments from \emph{Mrs.\ Doubtfire}, \emph{Big Bang Theory}, and \emph{Harry Potter}. A query about \emph{Mrs.\ Doubtfire} is localized to its corresponding interval, and the median frame of that interval is treated as the anchor. While coarse, this approach provides initial supervision for the ToMLoc retrieval module, giving the system a rough sense of \emph{when} to look for mental-state cues. Future work may refine this approximation by combining partial human labels or more advanced segmentation algorithms.
\vspace{-0.1cm}
\section{Experiments} \label{sec:experiments}
\vspace{-0.1cm}
In this section, we empirically assess the theory-of-mind (ToM) capabilities of video-based large language models (LLMs), focusing on two tasks: (1) \emph{frame localization}, which measures how well a model can identify the moment(s) revealing a character’s mental state, and (2) \emph{question answering} (QA), which tests whether the model can leverage these localized cues to make correct inferences.

\subsection{Experimental Setup}

We conducted experiments on the Social-IQ 2.0 dataset~\cite{siq2}, which comprises 1{,}403 video clips from real-world YouTube interactions, paired with 8{,}076 questions. Each question offers four possible answers, only one of which is correct. These queries often involve social or emotional reasoning, making the dataset well-suited to testing ToM abilities.

\paragraph{Preprocessing.}
Before finetuning, we normalized and resized all clips, then sampled up to 100 frames per video at uniform intervals. If a video exceeded 100 frames, we spaced the frame extraction more sparsely to maintain a manageable input size. We also ensured consistent resolution across clips, which helps standardize visual features.

\paragraph{Models Compared.}
We compared a baseline video LLM (Video-ChatGPT) trained on a general-purpose video dataset (ActivityNet~\cite{caba2015activitynet}) against two variants:
\begin{itemize}
    \item \textbf{Video-ChatGPT + ActivityNet/ToMLoc:} Finetuned on our ToM Localization (ToMLoc) task, aimed at retrieving frames containing crucial mental-state cues.
    \item \textbf{Leave-Frame-Out:} A data-augmentation approach that systematically omits certain frames during training, encouraging the model to identify essential frames more robustly.
\end{itemize}

\subsection{The Frame Localization Task}

We first evaluate the models on the frame localization task, which requires identifying the specific frames within a video that are most informative for inferring a character's mental state in response to a ToM-related query. This task moves beyond coarse activity recognition and demands sensitivity to subtle, temporally grounded social cues.

\begin{table*}[ht!]
\renewcommand{\arraystretch}{1.3}
\caption{Frame Localization Task Results on Social-IQ 2.0}
\label{table:frame_localization_results}
\centering
\begin{tabular}{l||l||c}
\hline
\bfseries Model & \bfseries Trained/Finetuned on & \bfseries Accuracy \\
\hline\hline
Video-ChatGPT  & ActivityNet                   & N/A             \\
\hline
Video-ChatGPT  & ActivityNet/ToMLoc            & 69.06\%           \\
\hline
Leave-Frame-Out & ActivityNet/ToMLoc                & 77.97\%           \\
\hline
\end{tabular}
\end{table*}

As shown in Table~\ref{table:frame_localization_results}, the baseline model pretrained only on ActivityNet~\cite{caba2015activitynet} is unable to localize relevant frames, reflecting its focus on broad activity categories rather than nuanced social inference. Finetuning on ToMLoc yields a substantial improvement, with accuracy rising to 69.06\%, demonstrating that explicit supervision for mental-state retrieval enables the model to attend to subtle cues such as facial expressions and gestures. The Leave-Frame-Out strategy achieves the highest accuracy at 77.97\%, suggesting that training the model to handle missing frames encourages it to more precisely identify and rely on the most diagnostically relevant moments for theory-of-mind reasoning.

\subsection{QA Performance on Social-IQ 2.0}

We next assess the models' ability to answer ToM-rich multiple-choice questions in Social-IQ 2.0, where success requires integrating spatiotemporal cues and social context to select the correct answer from four options.

\begin{table*}[ht!]
\renewcommand{\arraystretch}{1.3}
\caption{Finetuning on frame localization (ToMLoc) improves QA performance.}
\label{table:qa_performance}
\centering
\begin{tabular}{l||l||c}
\hline
\bfseries Model & \bfseries Trained/Finetuned on & \bfseries QA Accuracy \\
\hline\hline
Random Guesses           & N/A                           & 25.00\%  \\
\hline
Video-ChatGPT            & ActivityNet/Social-IQ 2.0 QA  & 39.61\%  \\
\hline
Video-ChatGPT            & ActivityNet/ToMLoc            & 47.15\%  \\
\hline
\end{tabular}
\end{table*}

As shown in Table~\ref{table:qa_performance}, a random-choice baseline achieves 25\% accuracy, as expected for a four-way multiple-choice task. Finetuning an ActivityNet-pretrained model on Social-IQ 2.0 increases performance to 39.61\%, indicating some ability to leverage social context but limited ToM inference. Further finetuning on ToMLoc yields a notable improvement to 47.15\%, underscoring the importance of grounding question answering in localized, temporally precise mental-state cues for robust theory-of-mind reasoning.

\subsection{Semi-Supervised Labels}

\begin{table*}[ht!]
\renewcommand{\arraystretch}{1.3}
\caption{Frame Localization on Semi-Supervised Labels is a Challenging Task}
\label{table:semi_supervised_labels}
\centering
\begin{tabular}{l||l||c}
\hline
\bfseries Model & \bfseries Trained/Finetuned on & \bfseries Accuracy (Strict) \\
\hline\hline
Random Guesses           & N/A                           & 1.00\%  \\
\hline
Video-ChatGPT            & ActivityNet                   & N/A     \\
\hline
Video-ChatGPT            & ActivityNet/ToMLoc            & 0.71\%  \\
\hline
\end{tabular}
\end{table*}

We also evaluate the use of semi-supervised labels generated by GPT-4V~\cite{openai2023gpt4v} as an alternative to manual annotation. As shown in Table~\ref{table:semi_supervised_labels}, models trained with these labels achieve only a marginal improvement over random guessing and remain far behind those trained with human-annotated data. This outcome indicates that, although GPT-4V can offer some supervision when manual labels are unavailable, current semi-supervised methods are insufficient for capturing the subtle social and mental-state cues necessary for accurate frame localization. These findings reinforce the critical role of high-quality human annotation in this domain and suggest that further progress in automated or hybrid human-AI labeling strategies will be essential to close the performance gap.

Our empirical results underscore how \emph{ToM-specific data} (via ToMLoc) substantially improves both frame localization and question answering (QA) performance. This section highlights key takeaways, along with limitations and avenues for future work.

\subsection{Finetuning and Model Performance}

Finetuning on ToMLoc data significantly enhances the model’s ability to detect critical social cues—such as facial expressions, gestures, and shifting emotional states—which are crucial for theory-of-mind (ToM) reasoning. As a result, both frame localization and QA tasks benefit, demonstrating that targeted data improves a model’s grasp of spatiotemporal context.

Nonetheless, real-world videos can feature complex challenges like occlusions, poor lighting, or overlapping dialogues. These issues can obscure vital cues and reduce model accuracy. Future work could explore diverse training sets, data augmentation (e.g., random cropping, brightness shifts), or specialized architectures for multi-agent interactions.

\subsection{Limitations and Future Directions}

While our results highlight the promise of ToM-specific data and targeted finetuning, several open research challenges remain. First, the scarcity of high-quality, diverse, and well-annotated video datasets continues to impede progress. The process of manually labeling frames or intervals that convey mental-state inferences is labor-intensive, and current semi-supervised approaches still fall short of expert human annotation. Addressing this bottleneck will require more sophisticated annotation protocols, scalable crowdsourcing strategies, or hierarchical labeling frameworks.

Another major challenge is generalization. Social norms, cultural cues, and subtle body language vary widely across contexts, making it difficult for models trained on a single dataset to transfer their reasoning skills to new domains. Future work should explore domain adaptation techniques and cross-cultural benchmarks to ensure robust ToM inference in diverse interpersonal settings.

Finally, modeling complex multi-character dynamics remains an open problem. Many social scenes involve multiple agents with distinct and evolving beliefs, and their mental states often influence one another over time. Capturing these interactions will likely require specialized modules for multi-agent belief tracking and memory mechanisms capable of handling longer, more intricate sequences.

\subsection{Summary of Findings}

In this work, we explored the potential for multimodal large language models (MLLMs) to exhibit emergent theory-of-mind (ToM) reasoning in video-based tasks. We introduced the Video Theory of Mind (VToM) architecture, which integrates textual and visual modalities to enhance ToM inference. Our pipeline, which learns to retrieve relevant frames and ground inference in spatiotemporal cues, significantly improves performance on social reasoning benchmarks.

Through the development of Theory of Mind Localization (ToMLoc), we proposed a new frame retrieval task that explicitly targets the localization of mental-state cues within video sequences. Our empirical results demonstrate that finetuning MLLMs on ToMLoc improves both frame localization and question-answering performance, validating the importance of grounding ToM reasoning in visual evidence.

Beyond the immediate technical contributions, our findings highlight the broader implications of video-based ToM reasoning in AI systems. Multimodal models capable of capturing human mental states could enable more natural and intuitive human-AI interactions. However, challenges remain, including dataset limitations, cultural variability in social reasoning, and the need for explainable spatiotemporal modeling.

Future work should focus on expanding video-based ToM benchmarks, refining annotation strategies for mental-state tracking, and novel architectures that explicitly model dynamic social interactions. These advancements may lead to \textit{human-like} AI systems that can reason about and respond to human emotions and intentions in complex, real-world scenarios.
\vspace{-0.1cm}
\section{Conclusion}
\vspace{-0.1cm}

Recent work has revealed that large language models (LLMs) can exhibit emergent theory-of-mind (ToM) capabilities—inferring human beliefs, desires, and intentions from text alone. Yet, everyday social reasoning often unfolds visually in dynamic contexts. This paper investigates whether multimodal LLMs can similarly demonstrate ToM skills in video-based tasks. Concretely, we propose a pipeline that fuses video and text signals, retrieves the most relevant frames for each query, and answers questions requiring spatio-temporal social understanding. We introduce a new frame localization benchmark, Theory of Mind Localization (ToMLoc), and show that finetuning a state-of-the-art Video-ChatGPT model on ToMLoc significantly improves performance on Social-IQ 2.0. Our results suggest that bridging textual and visual modalities is essential for capturing complex mental states in real-world scenarios. Moreover, retrieving key frames enhances interpretability by revealing how the model arrives at its inferences. These findings highlight the promise of video-based approaches for achieving more human-like social intelligence in LLMs.


\section*{Acknowledgment}
We thank Microsoft Accelerate Foundation Models Research
(AFMR) for their generous support, whose grant of Azure credits provided access to Azure OpenAI models for evaluation and semi-supervised labels. Additionally, Azure's GPU-accelerated compute resources were indispensable to finetuning large language models for our new tasks.

\bibliography{paper}
\bibliographystyle{assets/plainnat}

\newpage
\appendix
\onecolumn

\end{document}